%% file: CoG2019-for-arXiv.tex
\documentclass[conference]{IEEEtran}
\IEEEoverridecommandlockouts
\usepackage{cite}
\usepackage{amsmath,amssymb,amsfonts}	
\usepackage{algorithmic}
\usepackage{graphicx}
\usepackage{textcomp}
\usepackage{xcolor}

\usepackage{url}
\usepackage{bm}
\usepackage [english]{babel}	
\usepackage{xspace}
\usepackage{colortbl}		
\usepackage{tabularx}
\usepackage{subfigure}
\usepackage[draft]{changes}	

\usepackage{tikz}
\usetikzlibrary{shapes}
\usetikzlibrary{arrows}				
\usetikzlibrary{positioning}	
\usetikzlibrary{shadows}
\usetikzlibrary{calc}					
\usetikzlibrary{matrix} 	    

\usepackage{fancyhdr}
\def\BibTeX{{\rm B\kern-.05em{\sc i\kern-.025em b}\kern-.08em
    T\kern-.1667em\lower.7ex\hbox{E}\kern-.125emX}}

\pdfoutput=1									

\newcommand{\StateObservation}[1]{{\it StateObservation}\xspace}
\newcommand{\StateObsND}[1]{{\it StateObsNondeterministic}\xspace}
\newcommand{\GameBoard}[1]{{\it GameBoard}\xspace}
\newcommand{\PlayAgent}[1]{{\it PlayAgent}\xspace}
\newcommand{\AgentBase}[1]{\href{\#hrefPlayAgent}{AgentBase}\xspace}
\newcommand{\AgentState}[1]{\href{\#hrefAgentState}{AgentState}\xspace}
\newcommand{\Evaluator}[1]{{\it Evaluator}\xspace}
\newcommand{\Feature}[1]{{\it Feature}\xspace}
\newcommand{\Arena}[1]{{\it Arena}\xspace}
\newcommand{\ArenaTrain}[1]{{\it ArenaTrain}\xspace}
\newcommand{\XNTupleFuncs}[1]{{\it XNTupleFuncs}\xspace}

\begin{document}
%
\title{
	General Board Game Playing for Education and Research in Generic AI Game Learning\\
}

\author{\IEEEauthorblockN{Wolfgang Konen}
\IEEEauthorblockA{\textit{Computer Science Institute}\\
\textit{TH K{\"o}ln -- Cologne University of Applied Sciences}\\
Gummersbach, Germany\\
wolfgang.konen@th-koeln.de
}

}

\IEEEpubid{\begin{minipage}{\textwidth}\ \\[12pt]
978-1-7281-1884-0/19/\$31.00 \copyright 2019 IEEE \end{minipage}}

\maketitle

\begin{abstract}
We present a new \textbf{g}eneral \textbf{b}oard \textbf{g}ame (GBG) playing and learning framework. GBG defines the common interfaces for board games, game states and their AI agents. It allows one to run competitions of different agents on different games. It standardizes those parts of board game playing and learning that otherwise would be tedious and repetitive parts in coding. GBG is suitable for arbitrary 1-, 2-, $\ldots, N$-player board games. It makes a generic TD($\lambda$)-n-tuple agent for the first time available to arbitrary games. On various games, TD($\lambda$)-n-tuple is found to be superior to other generic agents like MCTS. GBG aims at the educational perspective, where it helps students to start faster in the area of game learning. GBG aims as well at the research perspective by collecting a growing set of games and AI agents to assess their strengths and generalization capabilities in meaningful competitions. Initial successful educational and research results are reported.
\end{abstract}




\section{Introduction} \label{sec:introduction}

\subsection{Motivation}
General board game (GBG) playing and learning 
is a fascinating area in the intersection of machine learning, artificial intelligence (AI) and game playing.
It is about how computers can learn to play games not by being programmed but by gathering experience and learning by themselves (self-play). Recently, AlphaGo Zero~\cite{silver2017AlphaGoZero} and AlphaZero~\cite{silver2017AlpaZeroChess} have shown remarkable successes for the games of Go and chess with high-end deep learning and stirred a broad interest for this subject.



A common problem in game playing is the fact, that each time a new game is tackled, the AI developer has to undergo the frustrating and tedious procedure to write adaptations of this game for all agent algorithms. Often he/she has to reprogram many aspects of the agent logic, only because the game logic is slightly different to previous games.
Or a new algorithm or AI is invented and in order to use this AI in different games, the developer has to program instantiations of this AI for each game. 

GBG is a recently developed software framework consisting of classes and interfaces which standardizes common processes in game playing and learning. If someone programs a new game, he/she has just to follow certain interfaces described in the GBG framework, and then can easily use and test \textit{all} AIs in the GBG library on that game. Likewise, it is easier to implement new AI agents.


 
The first motivation for GBG was to facilitate the entry for our computer science undergraduate and graduate students
into the area of game learning (\textit{educational perspective}): Students at our faculty are often very interested in game play and game learning. Yet the time span of a project, a bachelor's or a master's thesis is usually too short to tackle both game development and AI development 'from scratch'. The GBG framework allows a quicker start for the students, it offers over time a much larger variety of AIs and games and it allows students to focus on their specific game and AI research.

The second motivation is to facilitate for researchers the competition between AI agents on a variety of games to drive the research for versatile (general) AI agents (\textit{research perspective}). This has some resemblance to General Game Playing (GGP), but there are also important differences (see Sec.~\ref{sec:relatedWork} and \ref{sec:introGBG}).

The third motivation is to have a framework where AI agents learning by themselves (similar to AlphaZero) can be compared with strong- or perfect-playing agents for certain games.
This allows to answer the question: How close to perfect-playing performance comes a self-learning AI agent on game XYZ?

Last but not least, a competition with a large variety of agents and games helps to solve the often non-trivial question on how to evaluate the real strength of game-playing agents in a fair manner.  

The rest of this document is structured as follows: After a short summary of related work in Sec.~\ref{sec:relatedWork}, 
Sec.~\ref{sec:methods} gives an overview of GBG: classes, agents and games currently implemented in GBG. 
Sec.~\ref{sec:results} discusses first results obtained with this framework. Sec.~\ref{sec:conclusion} concludes.

\begin{table*}%
\caption{The main classes of GBG. 'Type' is either interface (IF) or abstract class (AC).
}
\label{tab:classes}
\begin{tabularx}{1.0\textwidth}{|l|l|X|}
\arrayrulecolor[gray]{0.0}
\hline\hline\rowcolor[gray]{0.9}
Name      					& Type & Description\\
\hline\hline\rowcolor[gray]{1.0}
\StateObservation\	&  IF & defines an abstract board game: state, rewards, available actions, game rules, ...			  \\
\StateObsND\	 			&  IF & derived from \StateObservation\:: additional methods for nondeterministic games  \\
\GameBoard\					&  IF & game board GUI: display states \& action values (inspection, game play), HCI for human players, ... \\
\PlayAgent\ 				&  IF & the agent interface: select next action, predict game value, perform training, ...\\
\Arena\							&  AC & meeting place for agents on an abstract game: load agents, play game, logs, evaluation, competition  \\
\ArenaTrain\				&  AC & derived from \Arena\ for additional capabilities: parametrize, train, inspect \& save agents\\
\hline
\hline
\end{tabularx}
\end{table*}

\subsection{Related Work}
\label{sec:relatedWork} 
One of the first general game-playing systems was Pell’s \textsc{Metagamer}~\cite{pell1996strategic}. It played a wide variety of simplified chess-like games.

Later, the discipline \textbf{General Game Playing (GGP)}~\cite{genesereth2014GGP,mandziuk2012generic} became a wider coverage and it has now a long tradition in artificial intelligence: Since 2005, an annual GGP competition organized by the Stanford Logic Group~\cite{genesereth2005GGPcompetition} is held at the AAAI conferences. Given the game rules written in the so-called \textit{Game Description Language} (\textit{GDL},~\cite{love2008GDL}), several AIs enter one or several competitions. As an example for GGP-related research, Mandziuk et al.~\cite{mandziuk2012generic} propose a universal method for constructing a heuristic evaluation function for any game playable in GGP. 
With the extension \textit{GDL-II}~\cite{thielscher2010general}, where \textit{II} stands for \glqq Incomplete Information\grqq, GGP is able to play games with incomplete information or nondeterministic elements as well.

GGP solves a tougher task than GBG: The GGP agents learn and act on previously unknown games, given just the abstract set of rules of the game. This is a fascinating endeavour in logical reasoning, where all informations about the game (game tactics, game symmetries and so on) are distilled from the set of rules at \textit{run time}. But, as Swiechowski~\cite{swiechowski2015recent} has pointed out, arising from this tougher task, there are currently a number of limitations or challenges in GGP which are hard to overcome within the GGP-framework: 
\begin{itemize}
	\item Simulations of games written in GDL are slow. This is because math expressions, basic arithmetic and loops are not part of the language.
	\item Games formulated in GDL have suboptimal performance as a price to pay for its universality: This is because \glqq it is almost impossible, in a general case, to detect what the game is about and which are its crucial, underpinning concepts.\grqq\ \cite{swiechowski2015recent}
	\item The use of Computational Intelligence (CI), most notably neural networks, deep learning and TD (temporal difference) learning, have not yet had much success in GGP. 
	As Swiechowski~\cite{swiechowski2015recent} writes: \glqq CI-based learning methods are often too slow even with specialized game engines. The lack of game-related features present in GDL also hampers application of many CI methods.\grqq\ Michulke and Thielscher~\cite{michulke2009neural,michulke2011IJCAI} presented first results on translating GDL rules to  neural networks and TD learning: Besides some successes they faced problems like overfitting, combinatorial explosion of input features and slowness of learning. 
\end{itemize} 

GBG aims at offering an alternative with respect to these limitations, as will be further exemplified in Sec.~\ref{sec:introGBG}. 
It has not the same universality as GGP, but 
agents from the CI-universum (TD, SARSA, deep learning, ...) can train and act fast on all available games.

Other works with relations to GBG: Kowalski et al~\cite{kowalski2015GGP} present so called \textit{Simplified Boardgames} where agents have material and position-piece features and their weights are trained by TD learning. These agents are compared with state-of-the-art GGP agents. -- A field related to GGP is
\textit{General Video Game Playing} (GVGP, \cite{levine2013GVGP}) which tackles video games instead of board games. Likewise, {$\mu$RTS} \cite{ontanon2015adversarial},\cite{barriga2017} is an educational framework for AI agent testing and competition in real-time strategy (RTS) games.
\textit{OpenAI Gym} \cite{brockman2016OpenAIgym} is a toolkit for reinforcement learning research which has also a board game environment supporting a (small) set of games. 

\section{Methods}
\label{sec:methods} 

\subsection{Introducing GBG}
\label{sec:introGBG} 
We define a \textbf{board game} as a game being played with a known number of players, $N=1,2,3,\ldots$, usually on a game board or on a table. The game proceeds through actions (moves) of each player in turn. This differentiates board games from video or RTS games where usually each player can take an action at any point in time. Note that our definition of board games includes (trick-taking) card games (like Poker, Skat, ...) as well. Board games for GBG may be deterministic or nondeterministic.

What differentiates GBG from GGP? -- GBG has not the same universality than GGP in the sense that GBG does not allow to present new, previously unknown games at \textit{run time}. However, virtually any board game can be added to GBG at \textit{compile time}. GBG then aims at overcoming the limitations of GGP as described in Sec.~\ref{sec:relatedWork}:
\begin{itemize}
	\item GBG allows fast game simulation due to the compiled game engine (10.000-90.000 moves per second for TD-n-tuple agents, see Sec.~\ref{sec:speedGBG}). 
	\item The game or AI implementer has the freedom to define game-related features or symmetries (see Sec.~\ref{sec:symmetries}) at compile time which she believes to be useful for her game.
	Symmetries can greatly speed up game learning. 
	\item GBG offers various CI agents, e.g. TD- and SARSA-agents and -- for the first time -- a \textbf{\textit{generic}} implementation of TD-n-tuple-agents (see Sec.~\ref{sec:agentOverview}), which can be trained fast and can take advantage of game-related features. With \textit{generic} we mean that the n-tuples are defined for arbitrary game boards (hexagonal, rectangular or other) 
and that the same agent can be applied to 1-, 2-, $\ldots$, $N$-player games.
	\item For evaluating the agent's strength in a certain game it is possible to include game-specific agents which are strong or perfect player for that game. Then the \textit{generic} agents (e.~g. MCTS or TD) can be tested against such specific agents in order to see how near or far from strong/perfect play the generic agents are on that game.\footnote{Note that in GGP agents are compared with other agents from the GGP league. A comparison with strong/perfect game-specific (non-GGP) agents is usually not made.} It is important to emphasize that the generic agents do \textit{not} have access to the specific agents during game reasoning or game learning, so they cannot extract game-specific knowledge from the other strong/perfect agents.
	\item Each game has a game-specific visualization and an inspect mode which allows to inspect in detail how the agents respond to certain game situations. This allows to get deeper insights where a certain agent performs well or where it has still remarkable deficiencies and what the likely reason is. 
\end{itemize}

GBG is written in Java and supports parallelization of multiple cores for time-consuming tasks. It is available as open source from GitHub\footnote{\url{https://github.com/WolfgangKonen/GBG}
	} and as such -- similar to GGP -- well-suited for educational and research purposes.

\subsection{Class and Interface Overview} 
\label{sec:classOverview}
A detailed description of the classes and interfaces of GBG is beyond the scope of this article. All classes and the underlying concepts are described in detail in a technical report elsewhere~\cite{Kone17a,Kone19a}. Here we give only a short overview of the most relevant classes in Table~\ref{tab:classes}.

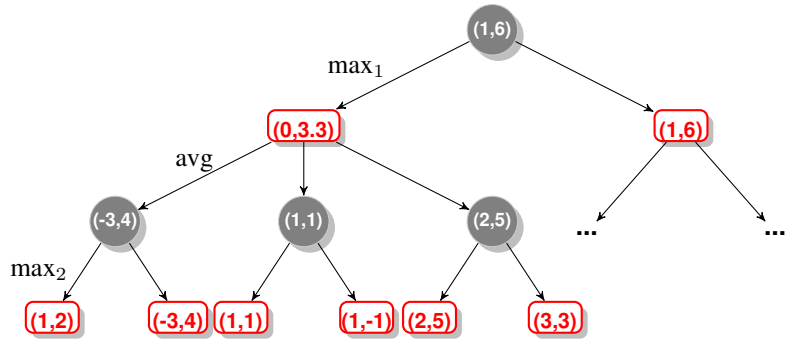
\begin{figure*}[!t]
\input{treeExpectimax}
\caption{Expectimax-N tree for $N$-player games. Expectimax-N is a generalization of Max-N \cite{korf1991maxN} for nondeterministic games. Shown is an example for $N=2$ and depth $d=3$. A node contains a tuple of game values for each player. The first level maximizes the tuple entry of the player to move (here: 1st player), the second level calculates the expectation value of all child nodes (grey circles), each having a certain probability of occurrence. The third level maximizes the tuple entry of the next player (here: 2nd player). At the leaves, the score tuple is obtained from the reward function of \StateObservation.. 
}
\label{fig:expectimaxTree}
\end{figure*}

\subsection{Agent Overview} 
\label{sec:agentOverview}

Table \ref{tab:agents} provides an overview of the agents currently available in GBG. Further agents are planned to be included. Agents are implemented in such a way that they are applicable to $N$-player games with arbitrary $N$. Therefore we do not use the Minimax algorithm, which is only for 2-player games, but its generalization Max-N as suggested by Korf~\cite{korf1991maxN}, which is viable for any $N$. 

Expectimax-N is the generalization of Max-N for nondeterministic games. Its principle is exemplified in Fig.~\ref{fig:expectimaxTree}. If we predict the score tuples in the leaves with any other agent, then
Max-N or Expectimax-N of depth $d$ becomes a generic $d$-ply look-ahead wrapper for that other agent. This can strengthen the quality of the other agent (at the expense of increased computational costs at play or competition time). Wrapper Max-N should be used for deterministic games, Expectimax-N for nondeterministic games. Thus, for each agent listed in Table~\ref{tab:agents} (except for HumanAgent, of course),  it is possible to construct further wrapped agents by wrapping them in a $d$-ply Max-N or Expectimax-N.

TD-n-tuple is the coupling of TD reinforcement learning with n-tuple features. It has been made popular by Lucas~\cite{Lucas08} with his success on Othello. An n-tuple is a set of $n$ board cells. If a cell has $L$ possible states, then there are $L^n$ n-tuple features, and each feature has a trainable weight. A network of n-tuples often has millions of weights. With reinforcement learning the network learns which of them are important. TD-n-tuple has optional temporal coherence learning~\cite{Beal99} integrated which facilitates learning through individual learning rates for each weight.

More information on the agents in Table \ref{tab:agents} is found in the relevant literature: MCTS~\cite{browne2012MCTS}, MCTS-Expectimax~\cite{Kutsch17}, TD~\cite{SuttBart98,Kone15c}, TD-n-tuple~\cite{Lucas08,Thil14,Bagh15,jaskowski2018mastering}, SarsaAgent~\cite{SuttBart98,Lucas08}.

\begin{table}%
\caption{Built-in agents of GBG. The first eight agents are AI agents. 
}
\label{tab:agents}
\begin{tabularx}{1.0\columnwidth}{|l|X|}
\arrayrulecolor[gray]{0.0}
\hline\hline\rowcolor[gray]{0.9}
Agent     			& Description\\
\hline\hline\rowcolor[gray]{1.0}
Max-N 					&  'Minimax' for $N$ player \cite{korf1991maxN}  \\
Expectimax-N 		&  Max-N extension for nondeterministic games  \\
MC					 		&  Monte Carlo (Sec.~\protect\ref{sec:2048}) \\
MCTS 						&  Monte Carlo Tree Search \cite{browne2012MCTS}\\
MCTS-Expectimax	&  MCTS extension for nondeterministic games  \\
TD		 					&  TD agent acc. to Sutton~\cite{SuttBart98} with user-supplied features  \\
TD-n-tuple 			&  TD agent with n-tuple features \cite{Lucas08} \\
SarsaAgent 			&  SARSA agent~\cite{SuttBart98} with n-tuple features \cite{Lucas08} \\
RandomAgent			&  Completely random move decisions \\
HumanAgent			&  A human user performs the moves \\
\hline
\hline
\end{tabularx}
\end{table}

\subsection{Competitions} 
\label{sec:competition}
A game competition is a contest between agents. Multiple objectives play a role in such a contest: 
(a)~the ability to win a game episode 
or a set of game episodes, maybe from different start positions or against different opponents (results clearly depend on the nature of the opponents); (b)~ability to win in several games; (c)~time needed during game play (either time per move or budget per episode or per tournament); (d)~time needed for training the agent. Other objectives may play a role in competitions as well.

Competition objectives may be combined, i.~e. how is the agent's ability to win if there is a constraint on the play time budget. 
When running this with different budgets, a curve 'win-rate vs. time budget' can be obtained.
-- Methods from multi-objective optimization (e.~g. Pareto dominance) can be used to compare and visualize competition results. 

GBG currently supports: 
\begin{itemize}
	\item pairwise, multi-episode competitions between agents, 
	\item multi-agent, multi-episode competitions (full round-robin tournaments or shortened tournaments) with optional Elo rating and Glicko rating included.
\end{itemize}

\subsection{Game Overview} 
\label{sec:gamesOverview}

The following games are currently available in GBG:
\begin{itemize}
	\item \textit{Tic-tac-toe}, a very simple 2-player game on a 3x3-board, mainly for test purposes.
	\item \textit{Nim}, a scalable 2-player game with $N$ heaps, each of arbitrary size. This game is strongly solved according to the theory of Bouton~\cite{bouton1901nim}. Yet some generic agents have difficulties in learning the right move for every situation. GBG allows to explore the possible reasons.
	\item \textit{2048}~\cite{cirulli2014} is a 1-player game on a 4x4-board with a high complexity: $(4\cdot 4)^{18}=4.7\cdot 10^{21}$ states. It is a game including chance events (nondeterministic game).
	\item \textit{Hex}, a scalable 2-player game on a diamond-shaped rectangular grid. It can be played on all sizes from 2x2, where it is trivial, up to arbitrary sizes, with 11x11 being a common size. 11x11 Hex has more legal states than chess and a higher branching factor. 
	\item \textit{Connect-4}, a 2-player game of medium-high complexity ($4.5\cdot 10^{12}$ states) for which a perfect-playing agent as evaluator is available. 
	\item \textit{Rubik's Cube}, the well known 1-player puzzle, either in the 2x2x2- or in the 3x3x3-version.
\end{itemize}

For the near future it is planned to include other games, especially those with  3 or more players which are not often covered in the game learning community. 

\subsection{Symmetries} 
\label{sec:symmetries}

Many board games exhibit symmetries, that is transformations of board states into equivalent board states with identical game value. For example, Tic-tac-toe and 2048 have eight symmetries (4 rotations $\times$ 2 mirror reflections). Hex (see Fig.~\ref{fig:5x5Hex}) has only one symmetry, the $180^o$-rotation. Game learning usually proceeds much faster, if symmetries are taken into account. GBG offers a generic interface to code symmetries and to use them during learning. 

\section{Results}
\label{sec:results} 

\subsection{The Educational Perspective}
\label{sec:educational} 

Here we report on the first educational progress made with the GBG framework. After the initial beta version of this framework was released, two computer science  students were interested in doing their bachelor's theses in this area~\cite{Kutsch17,Galitzki17}. The first student 
brought up the idea to solve the game 2048 with AI techniques (genetic algorithms, MCTS or similar). The second student started several months later and he was interested in tackling the scalable game Hex (the board size can be varied from 2x2 to 11x11 or even larger). Both worked enthusiastically on the topic and could generate first results within days or weeks. 

This would not have been possible without the GBG framework since developing and debugging a TD-n-tuple or MCTS agent is normally out of reach for a 6-12 weeks bachelor's thesis. Thanks to the framework, both (and other) agents were available and could be readily used for research and competitions. Similarly, the presence of a standardizing interface and the availability of sample agents made it easier for the students to develop variants and enhancements. In the case of 2048, the student developed the new agents MC and MCTS-Expectimax (see Sec.~\ref{sec:2048}). 

\begin{figure*}[tbp]
\centerline{\subfigure[Hex gameboard]{\includegraphics[width=0.40\textwidth]{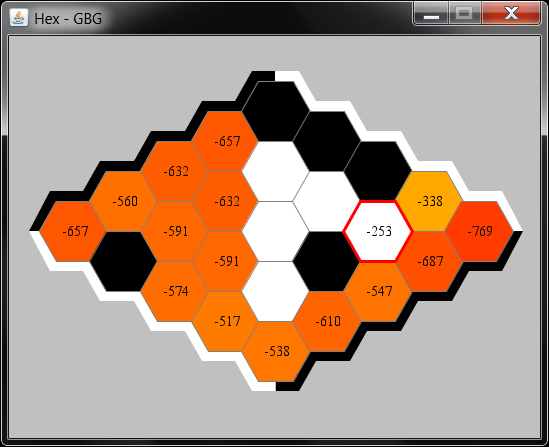} \label{fig:hexGameboard}}
\hfil
\subfigure[2048 gameboard]{\includegraphics[width=0.505\textwidth]{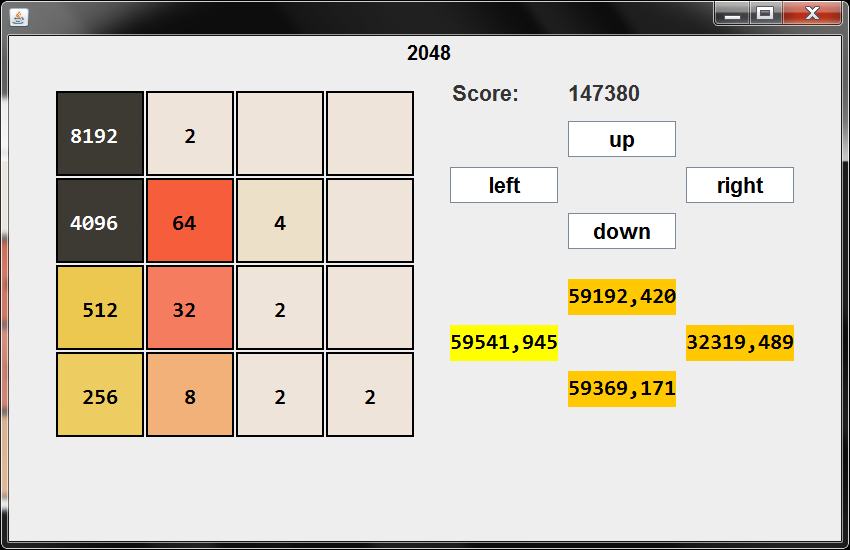} 
\label{fig:2048Gameboard}}}
\caption{(a) Hex gameboard example: The numbers and the color coding in the cells shows the agent's game values for the last move decision (White's turn). (b) 2048 gameboard example: The left part shows the 2048 playing field, the right part the possible actions. The numbers in the colored rectangles on the right show the agent's game values for the last move decision. The yellow rectangle indicates the direction of the last move which was 'left'. Note that 'right' has a drastically lower game value than all three other choices (it would destroy the column with the high tiles).
\label{fig:gameboards}} 
\end{figure*}


Finally, the resulting code is much better re-usable than code from individual projects since it is structured around well-defined interfaces. At the same time, using the interfaces for different games clarified some drawbacks in the initial beta-version design of GBG and led to improvements in interfaces and agents.

\subsection{Evaluation of Educational Benefits}
\label{sec:eval_edu}

In this subsection we evaluate the educational benefits of this project. Although at the time of writing this paper only two students had interacted with the GBG framework, we nevertheless tried to capture their opinion with the help of semi-structured interviews. [We note in passing that currently five more students and student teams, both graduate and undergraduate, are under way to conduct GBG projects.]

The interviews with the two students revealed the following facts: Both students fully agreed to the statement that GBG has accelerated their development process. They needed 1-2 days to get familiar with the framework and become productive. They felt well-supported by the documentation and the tutorials~\cite{Kone17a,Kone19a} available for GBG. Comparing the possibility to code their project \glqq from scratch\grqq\ or within the GBG framework, both voted clearly for GBG. They stated as positive elements of GBG: that it is easy for a programmer to implement new games and that the programmer does not need to dive into the algorithmic details of the agents in order to use them. They found the interfaces concise and clearly laid out. On their wishlist for GBG were more GUI elements to configure game-related settings (e.g. the size of a scalable game board).

Fig.~\ref{fig:gameboards} shows examples of gameboard GUIs developed by the students (Hex and 2048). The game learning for these two non-trivial games is only a first step, but it has delivered already some valuable insights. These research results are reported in the next three subsections.  

\begin{table}%
\caption{Results with GBG on the game 2048~\cite{Kutsch17}. Scores are averaged over 50 evaluation games. The results with TD-n-tuple were obtained after 200.000 training games. 0-ply refers to the plain TD-n-tuple agent, while the 2- and 4-ply results are achieved by wrapping the trained TD-n-tuple agent in an Expectimax-N agent of depth 2 and 4, resp.
\label{tab:results2048}
}
\centerline{
\begin{tabularx}{0.65\columnwidth}{|l|Xr|}
\arrayrulecolor[gray]{0.0}
\hline\hline\rowcolor[gray]{0.9}
Agent     			& & Average Score\\
\hline\hline\rowcolor[gray]{1.0}
%
	MCTS  & &  34.700	$\pm$ 4.100		\\
	MC		& &  51.500	$\pm$ 6.300		\\
	MCTSE	& &  57.000	$\pm$ 6.400		\\ \hline
	TD-n-tuple, 0 ply & &  131.000 $\pm$ 8.800\\
	TD-n-tuple, 2 ply & &  174.000 $\pm$ 6.600\\
	TD-n-tuple, 4 ply & &  196.000 $\pm$ 6.500\\ \hline\hline
\end{tabularx}
}
\end{table}

\subsection{Results 2048}
\label{sec:2048}


The 1-player game 2048~\cite{cirulli2014} is not easy to learn for computers, since a game episode can be rather long (several thousand moves). Initially, the game learning research started with two agents: MC and MCTS. The MC agent (repeated random rollouts for each action in a certain state) was meant as a simple baseline agent, while MCTS due to its tree structure was expected to act much better. But the comparison of these two agents led to two surprising results:
\begin{itemize}
	\item Both agents were not certain in the actions they advised: Repeating the rollouts on the same state often resulted in different actions suggested.
	\item MCTS was not superior to MC but instead inferior.
\end{itemize}
The first result comes from the nondeterministic nature of the game and triggered some research in the direction of agent ensembles. 
The reason for the second result is the nondeterministic nature of 2048 as well: If a plain MCTS is used, all next states resulting from a certain state-action pair (which differ by a random tile) are subsumed in \textit{one} node of MCTS. This node will then carry only one specific state, all further simulations will start from that state, and this leads to a severely limited exploration of the game tree.\footnote{MC on the other hand does not have this problem, since it does not store nodes.} Kutsch~\cite{Kutsch17} developed in response to this problem an MCTS-Expectimax (MCTSE) agent, where the tree is built according to MCTS principles, but the tree layers alternate between maximizing layers and expectation layers, similar to the Expectimax-N agent shown in Fig.~\ref{fig:expectimaxTree}.
The results in Table~\ref{tab:results2048} show that MCTSE is twice as good as MCTS and slightly better than MC.

First tests with the TD-n-tuple agent at the time of the bachelor's thesis were disappointing (scores below 30.000). The reason was that the implementation was not well-suited for nondeterministic 1-player games. This triggered a redesign of TD-n-tuple along the lines of Jaskowski~\cite{jaskowski2018mastering} (afterstates, new TD($\lambda$)-mechanism for long episodes). An afterstate is the game state after the agent has taken an action, but before the game environment has added the nondeterministic part. For game learning it is important to learn on afterstates and to average over chance events. The new version led to much better results, see Table~\ref{tab:results2048}: Scores between 131.000 and 196.000 are reached after 200.000 training games (6.6e8 learning actions).\footnote{Other parameters are: temporal coherence learning with 4 6-tuples, learning rate $\alpha=1.0$, $\lambda=\epsilon=0.0$}

Although Jaskowski~\cite{jaskowski2018mastering}, the current state-of-the-art for 2048, reports in the end largely better final scores up to 609.000, this is only achieved by additional methods designed specifically for 2048. Here we do not want to go into that direction, but are interested in the performance of a \textit{general purpose} TD-n-tuple agent. The comparison is only meant to be a correctness check that our implementation is comparable to \cite{jaskowski2018mastering}. 

We note in passing that Jaskowski~\cite{jaskowski2018mastering} reports for a plain TD(0.5)-agent (0-ply, with no 2048-specific extension) similar scores after 6.6e8 learning actions, even lower in the range of 80.000, where our agent achieves 131.000. 

\subsection{Results Hex}
\label{sec:Hex} 

Hex is a scalable 2-player game played on a board of variable size with hexagonal tiles and diamond shape. The goal for each player is to connect 'their' opposite sides of the board (see the black and white rims in Fig.~\ref{fig:gameboards}). 

\begin{figure}[tbp]%
\includegraphics[width=\columnwidth]{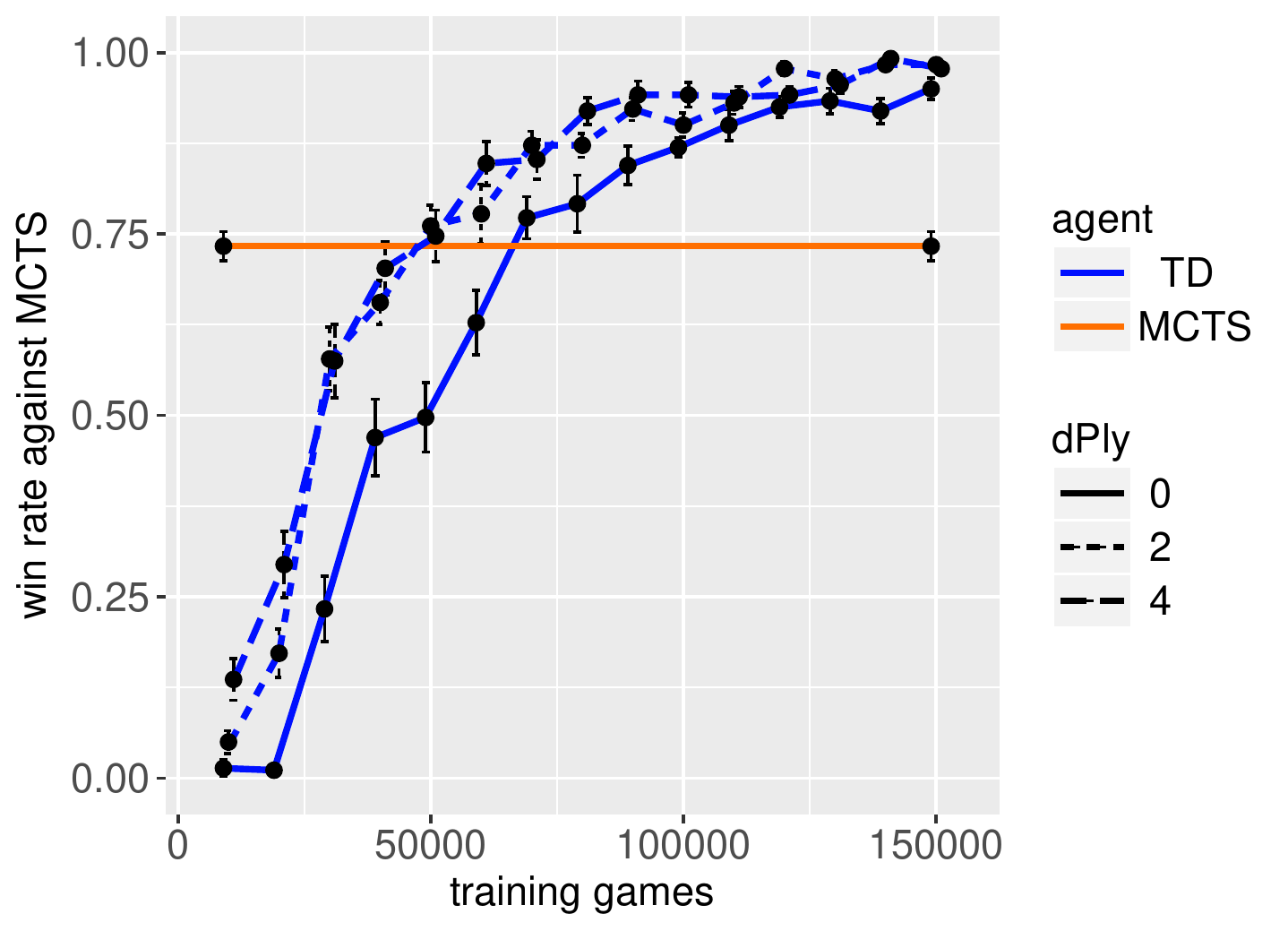}%
\caption{5x5 Hex: Training curves for TD-n-tuple agents with 25 random 6-tuples for various $d$-ply look-ahead wrappers. Shown as a function of training games is the win rate against MCTS for various starting boards where the agent can win (best is 1.0, average over 20 runs).
MCTS itself (horizontal line) reaches only a win rate of 0.74 on the same starting boards (average over 20 runs).}%
\label{fig:5x5Hex}%
\end{figure}

The bachelor's thesis~\cite{Galitzki17} conducted on Hex in the GBG framework was to the best of our knowledge the first application of a TD-n-tuple agent to the game of Hex. Other agents were tested as well. The main results are:
\begin{itemize}
	\item A TD agent with hand-designed features was successful for very small boards (2x2 and 3x3), but unsuccessful for all medium-size and larger boards (4x4 and up). 
	\item A general-purpose MCTS performs well for board sizes up to 5x5, but does not perform well on larger boards (6x6 and up).
	\item A TD-n-tuple agent with random n-tuples (automatic feature generation, no game knowledge required) was successful for \textit{all} board sizes from 2x2 up to 7x7. It could beat\footnote{We define \textit{'beat'} as winning from several starting position known to be a win for that agent.} the strong-playing computer program Hexy~\cite{anshelevich2002Hexy}. For board sizes 8x8 and higher it was so far not possible to construct a TD-n-tuple agent which would beat Hexy in a competition. 
\end{itemize}

The second result may come as a surprise, since it is well known that MoHex~\cite{Arneson2010MoHex}, a Hex playing agent based on MCTS, is a very strong playing Hex program that won several Hex Olympiads.  But MoHex incorporates many enhancements specific to Hex which are (not yet) generalizable to arbitrary board games. Our MCTS is a plain \textit{general-purpose} MCTS agent with no game-specific enhancements, and this has problems for larger Hex board sizes.

The last result is interesting from the generalization perspective: The TD-n-tuple agent, which was made popular by Lucas~\cite{Lucas08} with his success on Othello, later extended to TD($\lambda$) by Thill~\cite{Thil14} and successfully applied to Connect-4~\cite{Bagh15}, was taken nearly unmodified in the GBG-framework and applied to Hex without incorporating any game-specific knowledge. This demonstrates that the TD-n-tuple approach nicely generalizes to other games. It is much easier to apply to new games than the TD agent with its need for user-defined features. 

Fig.~\ref{fig:5x5Hex} shows the training performance of a TD-n-tuple agent on 5x5 Hex. There is a small but significant improvement when wrapping the plain TD agent (0-ply) in a 2-ply look-ahead with Max-N. Going from 2-ply to 4-ply look-ahead gives no further improvement. TD-n-tuple outperforms MCTS.\footnote{MCTS parameters: 10.000 iterations, depth 10, $K_{UCT}=1.414$.} 


\begin{figure}[tbp]%
\includegraphics[width=\columnwidth]{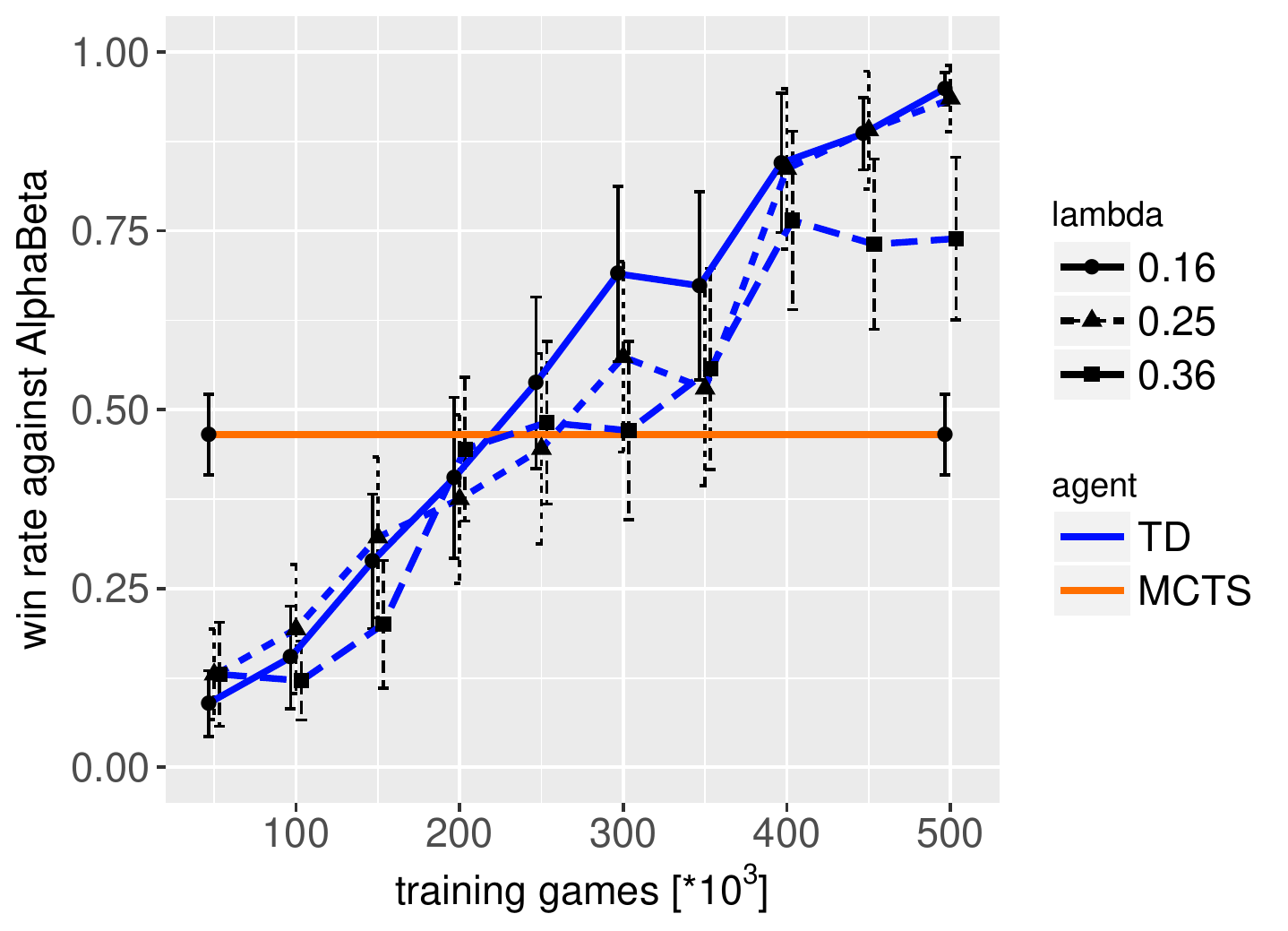}%
\caption{Connect-4: Training curves for TD($\lambda$)-n-tuple agents with 70 8-tuples for various values of $\lambda$. Shown as a function of training games is the win rate against opponent AlphaBeta (perfect player) from the default starting board (best is 1.0, average over 3 runs).
MCTS (horizontal line) wins only less than half of the episodes against AlphaBeta.}%
\label{fig:Connect4}%
\end{figure}

\subsection{Results Connect-4}
\label{sec:Connect4} 

The previous results of our group on Connect-4~\cite{Thil14,Bagh15} could be successfully ported to the GBG framework: Connect-4 is a non-trivial board game of medium-high complexity for which we constructed a fast and perfect playing AlphaBeta (AB) agent. This game-specific agent was ported to GBG as well. 

GBG allows to train generic TD($\lambda$)-n-tuple agents which learn solely by self-play and reach near-perfect playing strength against the AlphaBeta agent: Fig.~\ref{fig:Connect4} shows the win rate \glqq TD($\lambda$)-n-tuple vs. AB\grqq\ for start positions where TD($\lambda$)-n-tuple can win. For $\lambda \leq 0.25$, an average win rate above 93\% is reliably reached after 500.000 training games.\footnote{Other parameters as in~\cite{Bagh15}: temporal coherence learning with exponential transfer function, $\beta=2.7$. Learning rate $\alpha=5.0$, $\epsilon_{\mbox{\scriptsize init}}=0.1$, $\epsilon_{\mbox{\scriptsize final}}=0.0$. Eligibility traces up to horizon cut $\lambda^n \geq 0.01$.} -- A plain MCTS (without game-specific enhancements) cannot master the game Connect-4.\footnote{MCTS parameters: 10.000 iterations, depth 10, $K_{UCT}=1.414$.}

\subsection{Speed of Agents in GBG}
\label{sec:speedGBG} 

For most CI agents it is desirable that they can perform many moves per second, since the training of CI agents often requires millions of learning actions and a fast game play allows agents to be wrapped in $d$-ply look-ahead without too long execution times. 

We show in Table~\ref{tab:movesPerSecond} the moves/second obtained in the GBG framework for various agents and games on a single core Intel i7-4712HQ (2.3~GHz). Since a Connect-4 episode lasts at most 42 plies, at least 7.900/42 = 188 episodes/second can be trained and 
40.400/42 = 961 episodes/second can be played. 

Furthermore, GBG allows to parallelize time-consuming tasks on multiples cores, if the tasks are parallelizable. For example the evaluation of a TD-n-tuple object is parallelizable over episodes, since evaluation does not modify the TD-n-tuple object. On the other hand, the evaluation of a single MCTS object is not parallelizable, because each MCTS search modifies the internal tree. To parallelize MCTS evaluation, each parallel thread has to have its own MCTS object. The speed-up of parallelizable tasks is nearly linear, so that for 6 cores about 500.000 moves/second can be reached in 2048 game play, which is quite fast.

\begin{table}%
\caption{Moves/second for various agents and games on a single core. $d$-ply refers to the wrapper depth during game play, 0-ply means unwrapped agent. Game learning is always done on unwrapped agents. }
\label{tab:movesPerSecond}
\begin{tabularx}{1.0\columnwidth}{|X|l||r|r|}
\arrayrulecolor[gray]{0.0}
\hline\hline\rowcolor[gray]{0.9}
Agent     				& Game 			& \multicolumn{2}{c|}{ Moves/second during ...}\\
\cline{3-4}\rowcolor[gray]{0.9}
									&      			& ... game learning & ... game play \\
\hline\hline\rowcolor[gray]{1.0}
TD-n-tuple [0-ply]& 2048 			&  66.000 & 94.000 \\
TD-n-tuple [2-ply]& 2048 			&  66.000 &  5.000 \\
MCTSE	(1000 iter)	& 2048 			&      -- &    120 \\ \hline
TD-n-tuple [0-ply]& Connect-4 &   7.900 & 40.400 \\
TD-n-tuple [2-ply]& Connect-4 &   7.900 &  5.100 \\
MCTS (1000 iter)	& Connect-4 &      -- &     54 \\ \hline
TD-n-tuple [0-ply]& 5x5 Hex		&  17.600 & 20.500 \\
TD-n-tuple [2-ply]& 5x5 Hex		&  17.600 &    700 \\
MCTSE	(1000 iter)	& 5x5 Hex		&      -- &     31 \\ 
\hline
\hline
\end{tabularx}
\end{table}

\section{Conclusion}
\label{sec:conclusion} 

We presented with GBG a new framework for general board game playing and learning. The motivation for this framework came initially from the educational perspective: to facilitate the first steps for students starting into the area of game learning, with sophisticated agents and with mechanisms for competition and comparison between agents and over games. 

We reported on first results obtained by students using GBG, which are encouraging from the \textit{educational perspective}: The students were able to integrate sophisticated agents into their game research and they could generate new research results within the short time span of their thesis projects. One student could contribute new agent variants (MC and MCTSE) to the GBG framework. Since it is possible to play the games and visualize results in various forms, it is easier to gain insights on what the agents have learned and where they have deficiencies.

The following results are interesting from the \textit{research perspective}: 
\begin{itemize}
	\item To be successful with nondeterministic games (like 2048) it is important to have appropriate nondeterministic structures in the agents as well: These are Expectimax-N (in contrast to Max-N), the Expectimax layers in MCTSE (in contrast to MCTS) and the afterstate mechanism~\cite{jaskowski2018mastering} in the TD-n-tuple agents.
	\item The general-purpose TD-n-tuple agent is successful in a quite diverse set of games: 2048, Connect-4, and the scalable game Hex for various board sizes from 2x2 to 7x7. The TD-ntuple agent can be taken \glqq off-the-shelf\grqq\ and delivers very good results.
	\item By \textit{successful} we mean in the case of Hex or Connect-4 win rates $\geq 90\%$ against perfect-playing opponents\footnote{not only from the default start position but also from other 1-ply start positions which are possible wins}.
	\item It is still an open research question how to advise the best possible n-tuple structure for larger Hex boards and whether it is possible to train such agents for 8x8 and larger boards solely by self-play.
	\item Plain MCTS(E) agents with no game-specific enhancements and only random (non-biased) rollouts were on all tested games inferior in quality and slower in game-play than TD-n-tuple agents\footnote{however, TD-n-tuple agents need training which MCTS(E) does not}. 
\end{itemize}
 
The aim of GBG is not to provide world-championship AI agents for highly complex games like Go or chess. This will be probably the realm of sophisticated deep learning approaches like AlphaGo Zero~\cite{silver2017AlphaGoZero} or similar. The aim of GBG is to study agents and their interaction 
on a variety of games with medium-high complexity. Given this game complexity, results can be obtained reasonably fast on cheap hardware. The agents and their algorithms are open, easily accessible and easily modifiable by students and researchers in the field of game learning. Yet the variety of games is complex enough to make the challenge for the agents far from being trivial.

It is the hope that GBG framework helps to attract students to the fascinating area of game learning and that it helps researchers in game learning to quickly test their new ideas or to examine how well their AI agents generalize on a large variety of board games.

\subsection{Future Work}
\label{sec:future}

The results presented in this paper are only first steps with the GBG framework. More games need to be implemented to assess the real generalization capabilities of agents. More agents and more elements to aid agent competition (Sec.~\ref{sec:competition}) are needed as well. 

A special focus will be set on contributing a variety of $N$-player games with $N>2$. Many agents available so far have been tested only on 1- or 2-player games. Agents like Max-N, Expectimax-N, TD and TD-n-tuple are implemented in such a way that they should generalize well to $N>2$, but this needs to be proven on $(N>2)$-games.  


\section*{Acknowledgment}

The author would like to thank Johannes Kutsch, Felix Barsnick and Kevin Galitzki for their contributions to the GBG framework.


%

\end{document}

%% file: treeExpectimax.tex
\centering
\begin{tikzpicture}[->,>=stealth',level/.style={sibling distance = 5.9cm/#1,  level distance = 1.5cm},scale=.85, align=flush center] 
\tikzset{
  treenode/.style = {align=center, inner sep=0pt, text centered,  sibling distance = 4cm, level distance = 1.0cm,
    font=\sffamily},
  ellipsis/.style={rectangle, draw=none, rounded corners=1mm, fill=none, level distance=1mm,
    text centered, anchor=north, text=black, font=\sffamily\bfseries},
  avgN/.style = {treenode, circle, white, font=\sffamily\bfseries\scriptsize, draw=black!20,drop shadow, 
    fill=black!50, text width=1.8em},    
  maxN/.style = {treenode, rectangle, rounded corners=1mm, red, draw=red, font=\sffamily\bfseries\footnotesize, drop shadow, 
    fill=white, text width=2.0em, text height=1.0em, text centered, thick}       		
}
\node [avgN] {(1,6) }
	child{ node(q1) [maxN, text width=2.7em] {(0,3.3) }
    child{ node(q2) [avgN] {(-3,4)} 
			child{ node(q3) [maxN] {(1,2)} 
      }
      child{ node [maxN, text width=2.2em] { (-3,4)}
      }   
    }
    child{ node [avgN] {(1,1)}
			child{ node [maxN] {(1,1)} 
			}
			child{ node [maxN] {(1,-1)} 
			}
		}
    child{ node [avgN] {(2,5)}
			child{ node [maxN] {(2,5)} 
			}
			child{ node [maxN] {(3,3)} 
			}
		}
	}
	child{ node [maxN] {(1,6) }
    child{ node [ellipsis] {...}
		}
    child{ node [ellipsis] {...}
		}
	}
; 
\node[above = 0.3cm of q1, draw=none] {\qquad \qquad $\mbox{max}_1$};  
\node[above = 0.2cm of q2, draw=none] {\qquad \qquad \qquad $\mbox{avg}$};  
\node[above = 0.15cm of q3, draw=none] {$\mbox{max}_2$\quad};  
\end{tikzpicture}